\pdfoutput=1
\documentclass[11pt]{article}

\usepackage[final]{coling}

\usepackage{times}
\usepackage{latexsym}

\usepackage[T1]{fontenc}
\usepackage[utf8]{inputenc}

\usepackage{microtype}

\usepackage{inconsolata}
\usepackage{multirow}
\usepackage{booktabs} 
\usepackage{graphicx} 
\usepackage{bbding}
\usepackage{algorithm}
\usepackage{algorithmic}
\usepackage{bbding}

\newcommand{\tabincell}[2]{\begin{tabular}{@{}#1@{}}#2\end{tabular}}

\usepackage{graphicx}

\title{LLM Sensitivity Evaluation Framework for Clinical Diagnosis}

\author{
\\
 \textbf{Chenwei Yan\textsuperscript{1,2}},
 \textbf{Xiangling Fu\textsuperscript{1,2,\footnotemark[2]}},
 \textbf{Yuxuan Xiong\textsuperscript{1,2}},
 \textbf{Tianyi Wang\textsuperscript{1,2}},
\\
  \textbf{Siu Cheung Hui\textsuperscript{3}},
 \textbf{Ji Wu\textsuperscript{4,5}},
  \textbf{Xien Liu\textsuperscript{4,\footnotemark[2]}}
\\
\textsuperscript{1}School of Computer Science, Beijing University of Posts and Telecommunications\\
 \textsuperscript{2}Key Laboratory of Trustworthy Distributed Computing and Service(BUPT), Ministry of Education\\
 \textsuperscript{3}Nanyang Technological University\\
 \textsuperscript{4}Department of Electronic Engineering, Tsinghua University
 \textsuperscript{5}College of AI, Tsinghua University
\\
\{chenwei.yan, fuxiangling, buptwty, bupt\_xyx\}@bupt.edu.cn \\
asschui@ntu.edu.sg, \{xeliu, wuji\_ee\}@mail.tsinghua.edu.cn 
}

\begin{document}
\maketitle

\renewcommand{\thefootnote}{\fnsymbol{footnote}}
\footnotetext[2]{Corresponding authors.}

\begin{abstract}
    Large language models (LLMs) have demonstrated impressive performance across various domains. However, for clinical diagnosis, higher expectations are required for LLM's reliability and sensitivity: thinking like physicians and remaining sensitive to key medical information that affects diagnostic reasoning, as subtle variations can lead to different diagnosis results. Yet, existing works focus mainly on investigating the sensitivity of LLMs to irrelevant context and overlook the importance of key information.
    In this paper, we investigate the sensitivity of LLMs, i.e. GPT-3.5, GPT-4, Gemini, Claude3 and LLaMA2-7b, to key medical information by introducing different perturbation strategies. 
    The evaluation results highlight the limitations of current LLMs in remaining sensitive to key medical information for diagnostic decision-making. The evolution of LLMs must focus on improving their reliability, enhancing their ability to be sensitive to key information, and effectively utilizing this information. These improvements will enhance human trust in LLMs and facilitate their practical application in real-world scenarios. Our code and dataset are available at https://github.com/chenwei23333/DiagnosisQA.

\end{abstract}

\section{Introduction}
    
\begin{figure}
        \centering
        \includegraphics[scale=0.35]{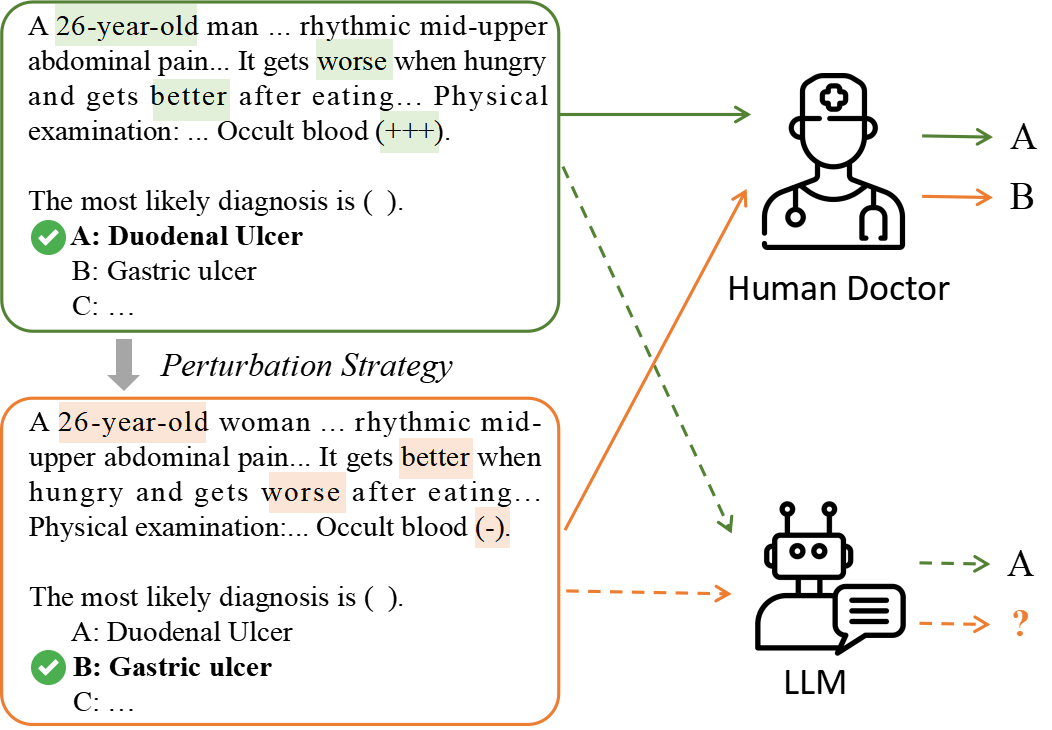}
        \caption{Human doctors are sensitive to key medical information. How do LLMs perform when key information is perturbed?}

        \label{fig:0}
    \end{figure}

    % LLM在医学评测中表现良好，但是在临床上使用仍有挑战：LLM是否可靠？
    Recently, Large Language Models (LLMs) have demonstrated their capabilities in various real-world tasks \cite{LLM-translation, LLM-coding}. In the medical domain, many kinds of evaluations have been proposed to assess LLMs' understanding of medical knowledge and their accuracy in medical licensing examinations. Although many LLMs have achieved impressive performance in these evaluations, even approaching the comprehension level of human doctors in specific tasks \cite{nature2023}, they have not yet been widely adopted in real clinical scenarios \cite{chung2024scaling}. One main reason is due to the reliability and interpretability of LLM-generated results.     
    As illustrated in Figure \ref{fig:0}, in the clinical diagnosis process, human doctors usually pay much attention to certain key information, such as the presence or absence of certain symptoms, and positive or negative results of certain tests. It is because subtle changes in these key information can lead to different diagnostic outcomes. These key medical information include gender, age, clinical symptoms, medical checkup results and so on. They provide essential information for doctors to reach a more accurate diagnosis. However, whether LLMs can behave as human doctors to maintain the same level of sensitivity to such key medical information, and provide an accurate and practical assessment and diagnosis is one of the critical factors in ensuring their reliability.
    
    % LLM具有敏感性，现有研究主要关注LLM对不重要信息的稳健性，但是忽略了LLM对重要信息的敏感性，尤其是在医疗领域。
    Previous works have shown that LLMs are sensitive to certain information. For example, \citet{pezeshkpour2023large} and \citet{zheng2023large} observed that LLMs are sensitive to the order of options in multiple-choice questions (MCQs), while \citet{shi2023large} found that LLMs can be easily distracted by irrelevant context. However, existing works focus mainly on studying LLMs' sensitivity to changes on insignificant data aspects such as the order of options in MCQs. 
    In real clinical practice, there are higher expectations for the sensitivity of LLMs, i.e., LLMs need to maintain a high level of sensitivity to key medical information as human doctors. When key medical information changes, LLMs should be able to detect it and ensure their responses remain sensible. This is crucial for enhancing their reliability and increasing human trust in LLMs.
    Currently, there is a lack of research on studying the sensitivity of LLMs to key medical information from a clinical perspective for evaluating the capabilities and limitations of LLMs. 
    
    % (3) Our novelty 
    In this paper, we propose a framework, named LLMSemEval, to evaluate the sensitivity of LLMs to key medical information, and assess their reliability for the clinical diagnosis.
    % The proposed framework consists of four steps.
    % First, we introduce four types of key medical information and two perturbation strategies. 
    % Then, we construct a clinical diagnosis-related dataset and generate eight perturbed datasets for evaluation.
    % Next, we select five state-of-the-art LLMs, including GPT-3.5, GPT-4 \cite{GPT-4}, Gemini, Claude3 and LLaMA2-7b \cite{llama-7b} for testing.
    % Finally, we evaluate the results using multiple metrics to assess their performance and sensitivity. 
    % We generate a series of sensitivity-focused datasets based on the interaction of four types of key medical information and two perturbation strategies. 
    Then, we evaluate LLM's sensitivity by comparing their performance on the original and generated derived datasets. The results indicate that GPT-4 outperforms the other models, but there remains a gap in its sensitivity for real-world clinical application. 
    Our contributions are summarized as follows:

    \begin{figure*}
        \centering
        \includegraphics[scale=0.39]{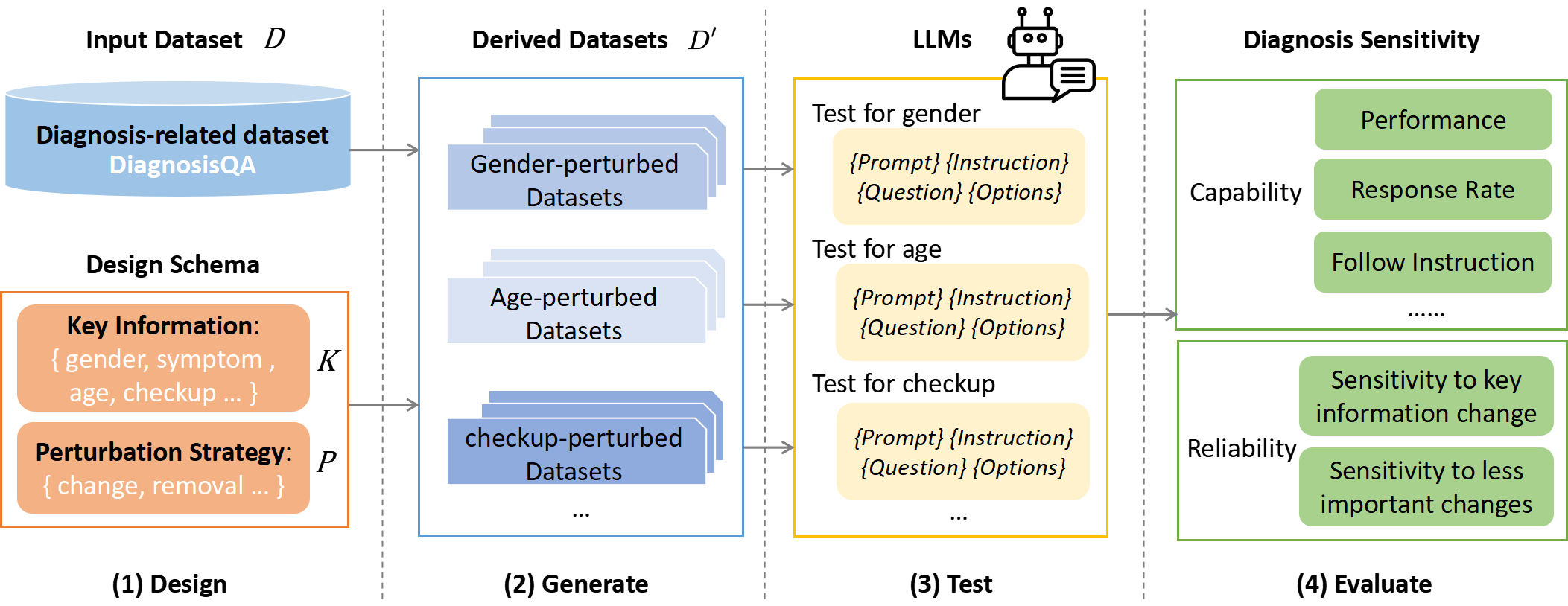}
        \caption{The proposed framework LLMSenEval for LLM sensitivity evaluation.}

        \label{fig:framework}
    \end{figure*}

    \begin{itemize}
    
        \item We propose a LLM sensitivity evaluation framework, named LLMSenEval, which provides a systematic approach for assessing the sensitivity of LLMs to key medical information in clinical diagnosis. To the best of our knowledge, this is the first work on proposing an evaluation framework to study the sensitivity of LLMs for clinical diagnosis.
    
        \item We propose a design schema for sensitivity evaluation, consisting of key information and perturbation strategy. We introduce four types of key medical information for clinical decision-making: age, gender, clinical symptoms, and checkup results. For each type of key medical information, we propose two perturbation strategies on change and removal.
        % 我们定义了设计敏感性测试的一个机制，由两部分组成，
        % We propose a design schema for sensitivity evaluation, consisting of two parts: key information and perturbation strategy.
        % In this schema,,
        
        \item Moreover, we construct a diagnosis-related QA dataset, DiagnosisQA, based on the publicly available MedQA dataset. By applying the proposed perturbation strategies, we generate a series of sensitivity-focused datasets for evaluation. The dataset has been publicly available on Github: https://github.com/chenwei23333/DiagnosisQA.
        
        % \item The experimental results demonstrate that all the LLMs under evaluation have limitations in sensitivity to medical key information, including the best-performing GPT-4, providing new direction for reliable medical LLMs.
        \item The experimental results reveal that all the evaluated LLMs, including the best-performing GPT-4, have limitations in their sensitivity to key medical information. These findings point to new directions for developing more reliable medical LLMs.

     \end{itemize}

\section{Related Work}
    % This section reviews the recent studies on the evaluation of the performance and sensitivity of LLMs in the medical domain. 
  
    \subsection{LLM Evaluation in the Medical Domain}
        The widespread adoption of LLMs \cite{brown2020, GPT-4, instructionGPT} has recently attracted much interest in evaluating their performance for various applications.
        For medical domain, 
        \citet{meng} proposed a biomedical knowledge probing benchmark MedLAMA, and evaluated the medical knowledge understanding of multiple pre-trained large language models.
        \citet{nature2023} tested the performance of LLM on factuality, comprehension, reasoning, possible harm and bias.
        \citet{zhou2024multifaceteval} investigated the degree and coverage of LLMs in encoding and mastering medical knowledge at four facets.
        % , namely comparison, rectification, discrimination and verification.

        In addition to evaluating their capabilities of understanding medical knowledge, some assessments are conducted to determine whether LLMs can pass medical licensing examinations and tackle real-world clinical problems. \citet{Valentin} conducted an evaluation using the US Medical Licensing Examination and found that GPT-3.5 reaches the passing score. Moreover, \citet{CN-benchmark2} proposed a Chinese medical examination benchmark, MedBench, for assessing the reasoning abilities of LLMs. \citet{cn-benchmark1} proposed CMExam, based on the Chinese National Medical Licensing Examination, and tested the performance of multiple general and medical domain-specific LLMs. However, although LLMs are able to achieve promising performance on many medical benchmarks, they still face many challenges for real-world applications, including issues on reliability and interpretability \cite{LLMprivacy, zhou2024multifaceteval, CN-benchmark2}, and there is a lack of research focusing on sensitivity in medical applications.

    \subsection{Sensitivity Evaluation of LLMs}
        Many studies and practices have observed the limitations in the sensitivity of LLMs. 
        In general, LLMs are sensitive to prompts, instructions and assigned roles to the extent that the inclusion of a sentence conveying reward or penalty in the prompt can affect their performance \cite{howtowriteprompts, xu2023earth, LLM-emotion}.
        
        Recent studies have shown that LLMs are sensitive to small changes in input context.
        \citet{pezeshkpour2023large} observed that LLMs are sensitive to the order of options in multiple-choice questions, and \citet{zheng2023large} pointed out that LLMs may exhibit a preference for specific options. Moreover, \citet{shi2023large} found that LLMs can be easily distracted by irrelevant context.
    
        Currently, the existing sensitivity studies mainly focus on how LLMs react to changes on insignificant data aspects. However, in clinical scenarios, apart from being undistracted by irrelevant information, a reliable LLM needs to be highly sensitive to important information. To our best knowledge, there is currently no related study on the evaluation of the sensitivity of LLMs to key medical information for clinical decision-making.

\section{Proposed Framework}
    In this section, we propose an evaluation framework for LLM sensitivity, called LLMSenEval. As illustrated in Figure \ref{fig:framework}, the LLMSenEval consists of four steps: Design, Generate, Test and Evaluate.
    \begin{table*}
        \centering
        \begin{tabular}{l|l|l|l|l}
            \toprule
            \multirow{2}{*}{\centering\tabincell{c}{Key Medical \\Information}} & \multirow{2}{*}{\centering \tabincell{c}{Perturbation \\Strategy}} & \multirow{2}{*}{Method} & \multirow{2}{*}{ Example} & \multirow{2}{*}{\tabincell{c}{Derived \\Dataset}}\\
            &&&&\\
            \midrule
            
            \multirow{2}{*}{Gender} & Gender Change & Swap the gender& female -> male &  $D_{GC}$ \\
                                    & Gender Removal & Use gender-neutral words& man -> patient&$D_{GR}$\\
            \hline
            \multirow{2}{*}{Age} &Age Change & Increase/decrease by 20\% &10->12& $D_{AC} $\\
                                 &Age Removal&Delete patient age& 2-year-old boy -> boy&$D_{AR}$\\
            \hline
            \multirow{2}{*}{Symptom} &Symptom Change& Swap presence/absence& with -> without&$D_{SC}$\\
                                     &Symptom Removal& Delete one clinical symptom&fever $\land$ cough -> cough&$D_{SR}$\\
            \hline
            \multirow{2}{*}{Checkup} &Results Change&Alter examination results&(+) -> (-)&$D_{CC}$\\
                                     &Results Removal&Delete one result&CT reveals... -> None&$D_{CR}$\\
            \bottomrule
        \end{tabular}
        \caption{Perturbation strategies on key medical information in the experiments.}
        \label{tab:perturbationStrategy}
    \end{table*}

    \subsection{Design}
        In the Design, we propose a design schema, consisting of key information and perturbation strategy. 
        % We identify four key medical information and select two perturbation strategies.
        
        \textbf{Key Information.} It is crucial to identify the required key information for clinical diagnosis.
        In the proposed framework, gender, age, symptoms and checkup results are identified as key medical information, denoted as $K=\{k_{gender}, k_{age}, k_{symptom}, k_{check}\}$.
        Gender and age are fundamental demographic characteristics in medical research, closely related to the occurrence and prognosis of numerous diseases. Clinical symptoms directly reflect a patient's physical condition, providing vital clues to narrow down the range of possible medical conditions of the patient. Combined with the medical checkup results, a comprehensive assessment and diagnosis can then be made. Overall, these key information provide important clues for doctors to make a correct diagnosis.

        % 是否要加上
        % We aim to evaluate whether the LLMs fully identify and utilize the key information, since they are vital for medical diagnosis. Based on this principle, we defined four types of key information after consulting with the clinical doctors and experts.
    
        \textbf{Perturbation Strategy.}
        Perturbing the key information is the basis for investigating the sensitivity of LLMs. We propose two perturbation strategies, namely key information change and key information removal, which are denoted as $P$. Based on the DiagnosisQA dataset, the details and examples of these strategies for perturbing each type of key medical information are shown in Table \ref{tab:perturbationStrategy}.
        
    \subsection{Generate}
        Given a clinical diagnosis dataset $D$ containing $n$ documents, denoted as $D=\{d_1, d_2, ..., d_n\}$, we extract the key information values for each document $d_i$ ($ 1 \le i \le n$) to generate a key-value dictionary $Info$ as follows:
            \begin{equation}
                Info = \{k: V_{k}\},
            \end{equation}
        where $k \in K$ is one of the four types of key medical information, and $V_{k}$ is the corresponding value of $k$ extracted from $d_i$.
        In our work, the values of key medical information are extracted using regular expression matching and keyword recognition \cite{negWord}.
    
        Then, we apply the perturbation strategies $P$ to the key medical information $K$.
        The document generated by perturbing the key information $k \in K$  using perturbation strategy $p \in P$ is denoted as $d_i^{kp} \in D_{kp}$.
        % This process is shown in Algorithm \ref{alg:1}.
        As a result, we obtain eight derived datasets: $D^{\prime} = \{D_{GC}, D_{GR}, D_{AC}, D_{AR}, D_{SC}, D_{SR}, D_{CC}, D_{CR}\}$ after applying the two perturbation strategies to the four types of key information.

    \subsection{Test}
        In this step, we evaluate the initial clinical diagnosis dataset $D$ and the derived datasets $D^{\prime}$.
        The input of LLMs is structured into prompt, instruction, question and options.
        
        \textbf{Prompt.} The input prompt includes the role (who you are), the task (what the input is) and the requirement (how to do). For clinical diagnosis, the prompt is written as \textit{"You are a medical expert. The following is a medical exam question. Please give the correct option and the corresponding explanation."}.
        
        \textbf{Instruction.} The instruction is given as \textit{"Return the result in JSON format with the following keys: Answer, Explanation."}, which specifies the return format of LLMs.

        \textbf{Question and Options.} The question and options are obtained from the QA question contents in $D$ and its derived datasets $D^{\prime}$.
    
    \subsection{Evaluate}

        \begin{figure*}
            \centering
            \includegraphics[scale=0.33]{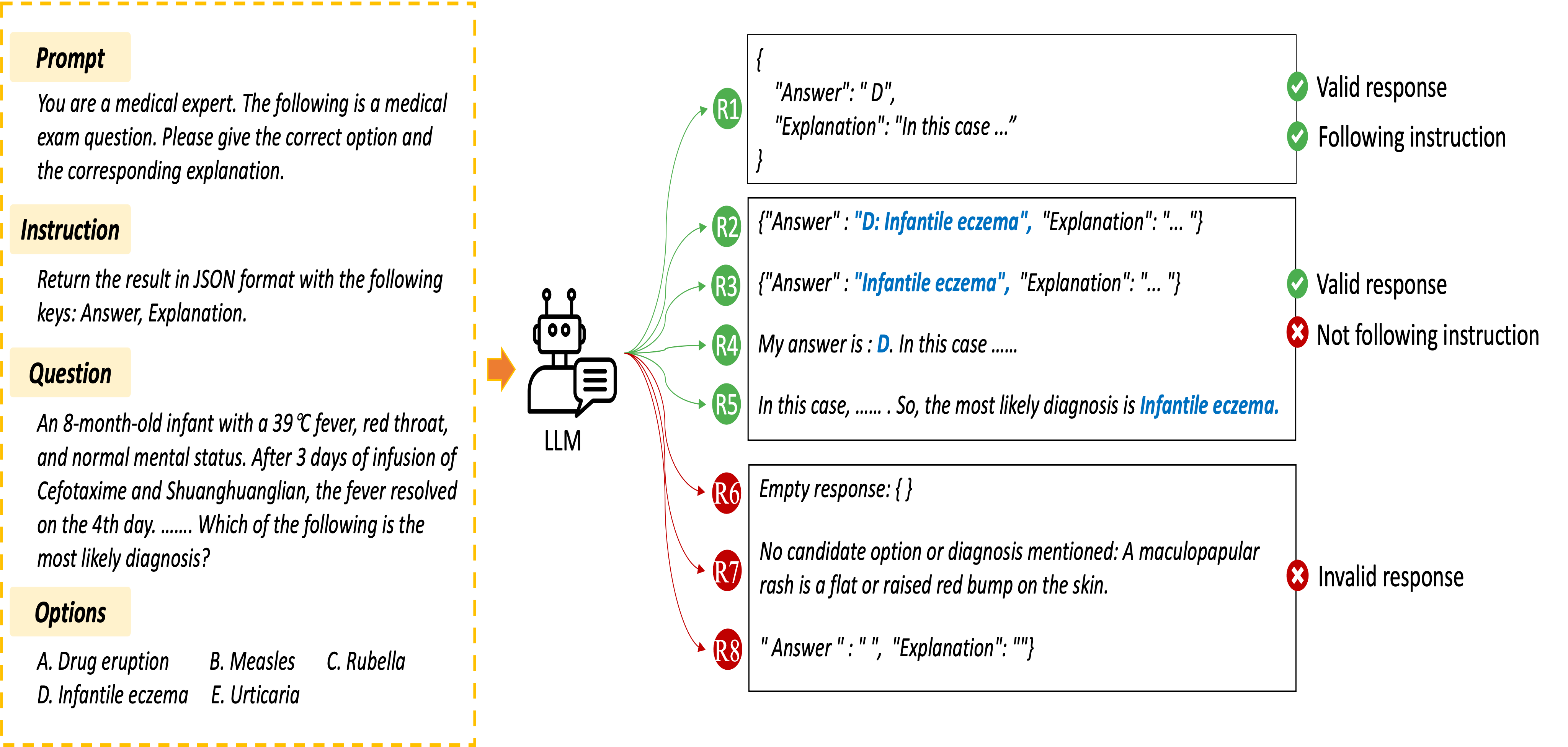}
            \caption{Valid and invalid responses from LLMs.}
            \label{fig:response}
        \end{figure*}
        
        The final step is to evaluate the responses using multiple metrics.
        
        \textbf{Evaluation Principle.} 
        % From the perspective of evaluation method, both rule-based automatic evaluation and human-based evaluation could be included.  
        The capabilities of LLMs can be evaluated from different perspectives. For example, comprehension and reasoning can assess their overall performance in understanding instructions and providing accurate predictions. Besides, sensitivity can be evaluated based on both changes in key information and less significant changes. A reliable LLM should be responsive to changes in key information while maintaining stability under unimportant variations.

        \textbf{Response Evaluation.} 
        The ideal response should be a JSON-formatted string with the following information: question, options, answer and explanation, each containing a non-empty value.
        However, some responses do not follow the given instructions, and they just return the answer in plain text format. Therefore, there is a need to preprocess the responses through regular expression matching to extract the answer in the required format.

        As illustrated in Figure \ref{fig:response}, a valid response is a non-empty response, with the answer option chosen from the candidate options. 
        In contrast, an invalid response occurs when LLMs refuse to provide a response, resulting in the "Answer" and "Explanation" fields being set to null. Additionally, if the responses do not include any options or diagnoses that match the candidate options, the values of "Answer" and "Explanation" are also set to null.
        When computing the metrics, such as accuracy, both valid and invalid responses are included, and empty responses are considered as wrong answers.

\section{Experiments}
    In this section, we discuss the experiments on evaluating the sensitivity of LLMs for clinical diagnosis based on the proposed LLMSenEval framework.
    We first adopt the four types of key medical information discussed in the framework and propose two perturbation strategies.
    We then construct the DiagnosisQA dataset and generate the derived datasets for evaluation. 
    After that, we select the LLMs for testing and define the metrics for evaluating the models' performance.

    \subsection{Dataset Generation}\label{section-dataset}
        In the experiments, we construct a diagnosis-related QA dataset based on the MedQA dataset \cite{MedQA}, by filtering out QA questions related to fundamental concepts and retaining those on case-based scenarios. Each case-based QA question provides details about patient symptoms, and checkup results, and concludes with "What is the most likely diagnosis?".

        \begin{table}[t]
            \centering
            \setlength{\tabcolsep}{1.5mm}{
            \begin{tabular}{l|rcc}
                \toprule
                    Dataset&\# Total & \# 5-Option & \# 4-Option  \\
                \midrule    
                     DiagnosisQA& 4,603 & 3,696 & 907 \\
                \midrule
                     $D_{GC}$, $D_{GR}$ &3,965 &3,236&729\\
                     $D_{AC}$,  $D_{AR}$ &4,008&3,342&666\\
                     $D_{SC}$, $D_{SR}$ &3,463&2,731&732\\
                    $D_{CC}$, $D_{CR}$ &3,439&2,903&536\\
                \bottomrule
            \end{tabular}}
            \caption{Statistics of the DiagnosisQA dataset and eight derived datasets.}
            \label{tab:dataset}
        \end{table}
        
        As a result, we obtain a dataset, named \textit{DiagnosisQA} dataset, comprising 4,603 QA questions. 
        Each question is accompanied by four or five candidate options.
        After applying perturbation strategies on DiagnosisQA, we obtain eight derived datasets. 
        The QA questions that could not be perturbed due to missing key information are not included in the derived datasets. The statistics of the derived datasets are shown in Table \ref{tab:dataset}.

        Furthermore, we annotate the derived datasets with the help of four professional physicians and experts, confirming that the answers to each question in these datasets have been reviewed and corrected by human doctors, thereby ensuring the correctness and testability of the QA questions. 

    \begin{table*}[t]
        \centering
        \setlength{\tabcolsep}{5mm}{
        \begin{tabular}{l|ccccc}
            \toprule
            Metrics& GPT-3.5 &GPT-4 &Gemini & LLaMA2-7b  &Claude3     \\
            \midrule
            Accuracy&61.38&\textbf{78.95}& 64.39&24.74&\underline{65.37} \\
            Precision&58.32&\textbf{68.78}&\underline{62.00}&28.00&58.80 \\ 
            Recall&54.69&\textbf{65.43}& \underline{53.74}&18.89&53.67\\
            F1-score&55.14&\textbf{67.03}&\underline{57.31}&16.44&55.92 \\ 
            \midrule
            RR&\textbf{99.95}&96.00&96.83&\underline{97.84}&94.92\\
            FIR&\underline{98.92}&\textbf{99.98}&68.43&0&94.03 \\
            
            \bottomrule
        \end{tabular}}
        \caption{Performances results (\%) of LLMs on the DiagnosisQA dataset without any perturbations. The results are obtained from an average of two trials. The best scores are in \textbf{bold}, and the second-best scores are \underline{underlined}.} 
        \label{tab:mainResults}
    \end{table*}
     % and the results of other LLMs are obtained from a one-off experiment

    \subsection{Selection of LLMs for Testing}
        We select five state-of-the-art LLMs for evaluation. From the GPT series, we choose the extensively evaluated versions: GPT-3.5-Turbo-0613 and the latest GPT-4-Turbo-2024-04-09. From the LLaMA series, we select LLaMA2-7b-chat \cite{llama-7b}. Additionally, we also include Google's Gemini-Pro and Anthropic's Claude-3-haiku-20240307 for testing in the experiments.

        For LLaMA2-7b-chat, we perform local deployment and inference. For the other models, we use APIs to invoke the service. All the model parameters are set with temperature of 0 and token limit of 1024. All the questions from the DiagnosisQA dataset and derived datasets are used for testing in the experiments.

    \subsection{Evaluation Metrics}\label{sec:metrics}
        In the experiments, we evaluate the LLM's performance in their capability and sensitivity. To do this, we use four standard evaluation metrics, namely accuracy, macro precision, macro recall and macro F1-score. Among them, accuracy is used as the primary metric for sensitivity evaluation.
        
        Moreover, we also introduce two additional metrics to better evaluate the LLM's capability on its instruction understanding and instruction following. 
        The first metric is the response rate (RR), which measures the number of valid responses of the model.
        The second metric is the followed-instruction rate (FIR), which evaluates how well the model follows the given instructions. These two metrics are defined as:
            \begin{equation}
                RR = \frac{\# validR}{N},
            \end{equation}
            \begin{equation}
                FIR = \frac{\# followedInstructionR}{\# validR},
            \end{equation}
        where $N$ is the total number of responses, $\# validR$ is the number of valid responses, and $\# followedInstructionR$ is the number of valid responses that follows the instructions.

\section{Performance Analysis on DiagnosisQA}
    Table \ref{tab:mainResults} shows the overall performance results of LLMs on DiagnosisQA dataset without any perturbations.
    % 基于DiagnosisQA上的分析
    % \subsection{Performance on Medical Diagnosis Task} 
        GPT-4 exhibits the highest accuracy at 78.95\%, significantly outperforming the second-ranked Claude3 by 13.58\%, highlighting its superior performance on medical diagnosis tasks. Furthermore, GPT-4 also demonstrates robust results in precision, recall and F1-score with 68.78\%, 65.43\% and 67.03\%, respectively, all of which also surpass the other models, indicating a well-balanced capability in both precision and recall.
        In contrast, LLaMA2-7b, likely due to its smaller parameter size, achieves much lower scores, with 24.74\% in accuracy. 
        Other models, such as Gemini and GPT-3.5, show moderate performance, with Gemini achieving an accuracy of 64.39\% and GPT-3.5 reaching 61.38\%.      
    
     % \subsection{Assessment of LLM Capacities}

        In addition to these metrics, Table \ref{tab:mainResults} also shows the performance results in RR and FIR. 
        All models show high response rate, exceeding 94\%.
        % The response was filtered due to the prompt triggering Azure OpenAI's content management policy
        However, in terms of adherence to the given instructions, GPT-4 demonstrates an FIR of 99.98\%, the highest among the five models. Conversely, LLaMA2-7b scores 0\% in FIR, indicating its responses require more post-processing, as it fails to generate the instructed formatted outputs.

        Overall, GPT-4 demonstrates the best overall performance based on DiagnosisQA, with its significantly higher accuracy and strict adherence to specified formats, showcasing its strong potential to serve as a medical AI assistant.

    % 基于不同语言子集上的分析
    % \subsection{Performance based on Different Language Subsets}\label{sec:subsets}
        
        % Overall, GPT-4-Turbo outperforms all other LLMs on accuracy, recall, F1-score, and FIR across three subsets. The performances in accuracy of the English, Simplified Chinese and Traditional Chinese subsets are 87.78\%, 75.46\%, and 80.82\%, respectively.
        
        % Apart from GPT-4-Turbo, other LLMs show varying levels of performance across different language subsets. On the English subset, GPT-3.5-Turbo outperforms other LLMs, achieving an accuracy of 71.95\%, while Claude3-haiku stands out on the Traditional Chinese subset with an accuracy of 69.68\%.
        % On the Simplified Chinese subset, Gemini and Claude3-haiku show comparable performance. Claude3-haiku achieves an accuracy of 64.38\%, slightly higher than Gemini's 63.30\%. However, Gemini exhibits higher recall and F1-score compared to Claude3-haiku.

    % 如果能加上LLAMA-70b就可以对比分析了
    %     \textbf{Scale.} Affected by parameter size, the Llama2-7B-chat model demonstrates a moderate performance, lagging behind other models and showing a marginal improvement over random guessing.

\section{Sensitivity Analysis on Derived Datasets }

    To evaluate the sensitivity of LLMs, we conduct further experiments on eight derived datasets. 
    Each dataset, based on manual annotation, contains two categories of questions.
    The first category is called \textit{Same Answer Subset (SAS)}, where the correct answers to the QA questions remain unchanged after perturbations, as the diagnosis results are not affected by changes in specific key information. 
    The second category is called \textit{Different Answer Subset (DAS)}, where the correct answers are changed due to the influence of perturbations. In these questions, the correct answers are updated to other correct options. In addition, as some perturbed questions do not contain the correct answer option, we add an additional option "None of the above". Furthermore, some perturbed questions introduce logical inconsistency for medical diagnosis, for instance, after a gender swap, the patient's gender will no longer align with the organs or diseases mentioned in the question.
    As such, we add another option "The question contains inconsistency" to indicate such situations. 
    
    Overall, each QA question contains only one correct answer.
    Note that it is important to distinguish the difference between these two categories of questions.
    For example, a patient's gender may not affect the diagnosis of a common illness like cold, while a patient's symptoms or checkup results can significantly influence the diagnosis results. 
        
    Following this, we conduct experiments to investigate sensitivity from two perspectives.
    First, we evaluate whether LLMs have low sensitivity to questions in the \textit{Same Answer Subset (SAS)} by measuring the change in accuracy before and after perturbations. A smaller difference indicates lower sensitivity, as these changes do not affect the final diagnosis results. Ideally, the accuracy should remain stable in this subset, as the correct diagnosis is not affected by the perturbations.
    Second, we evaluate whether LLMs have high sensitivity to questions in the \textit{Different Answer Subset (DAS)} by measuring their accuracy. Higher accuracy indicates greater sensitivity to the changes, as the pertubations directly impact the correct answer.
  
    % In addition, to mitigate the impact of randomness in model responses on accuracy, we randomly select 100 questions from the derived datasets to analyze the standard derivations of the LLM performance in medical diagnosis.
    % We conduct five repeated experiments on each LLM model. The resulting standard errors (SE) of model accuracy range from 0.22\% to 0.49\%.

    \subsection{Sensitivity Analysis on Gender}  
   
     Table \ref{tab:gender} shows the analysis results on gender. After applying gender change and removal perturbations, most question answers remain unchanged, as reflected in the \textit{Same Answer Subset}. 
     For this subset, the accuracy of LLMs varies, with some LLMs showing improvements while others degrading the performance. 
     GPT-3.5 exhibits low sensitivity to gender changes and removals, as the $\Delta$ accuracy lies within its standard error of 0.49\%.  
     In contrast, GPT-4 shows the most significant changes among the models.  Its $\Delta$ accuracy increases by 1.45\% when gender is changed and by 1.98\% when the gender is removed.
     
     % suggesting that it treats gender as irrelevant context or even as noise.

     % Gender 
        \begin{table}[h]
            \centering
            \setlength{\tabcolsep}{0.8mm}{
            \begin{tabular}{l|c|cc|cc}
            \toprule
                \multicolumn{2}{c|}{Subset}& \multicolumn{2}{c|}{SAS}& \multicolumn{2}{c}{DAS}\\
            \midrule
                 \multicolumn{2}{c|}{\multirow{2}{*}{\tabincell{c}{Perturbation\\(Total)}}}&\small{Change}&\small{Removal}&\small{Change}&\small{Removal}\\

               \multicolumn{2}{c|}{} &(3,587)&(3,961)&(378)&(4)\\
            \midrule
                 Metric&SE& \multicolumn{2}{c|}{$\Delta$ accuracy} & \multicolumn{2}{c}{accuracy}\\
                 \midrule
                 GPT-3.5&$\pm$0.49&\textbf{-0.06}&\textbf{+0.22}&0&0\\
                 GPT-4&$\pm$0.22&+1.45&+1.98&\underline{32.01}&0\\
                 Gemini&$\pm$0.35&-1.31&-1.11&0.79&0\\
                 Claude3 &$\pm$0.22&\textbf{-0.06}&+0.33&1.32&0\\
                 LLaMA2 &$\pm$0.41&\textbf{+0.36}&+1.06&0&0\\
                 \bottomrule
            \end{tabular}
            }
            \caption{The sensitivity results on gender. $\Delta$ indicates the difference in accuracy (\%) before and after perturbations. The results on SAS that are within the SE range are highlighted in \textbf{bold}. The best results on DAS are \underline{underlined}.}
            \label{tab:gender}
        \end{table}

        For the \textit{Different Answer Subset}, where perturbations result in changes to the correct answers, GPT-4 displays exceptional sensitivity. GPT-4 is able to identify such changes, answering 121 questions from 378 questions correctly with an accuracy of 32.01\%, far surpassing other models such as Gemini and Claude3, which can only answer 3 and 5 questions correctly, respectively.

    \subsection{Sensitivity Analysis on Age}  
    
    Table \ref{tab:age} shows the analysis results on age. 
    For the \textit{Same Answer Subset}, all models consistently exhibit low sensitivity to both age changes and removals. Among them, the $\Delta$ accuracy of GPT-3.5 remains within the standard error range under both change and removal perturbations.
    Although the $\Delta$ accuracy of other models exceeds their standard error ranges, most deviations are below 0.5\%, except in the case of age removal, where Gemini shows a decrease of 0.85\% in $\Delta$ accuracy and GPT-4 experiences a decline of 1.2\%.

        % age
        \begin{table}[t]
            \centering
            \setlength{\tabcolsep}{0.8mm}{
            \begin{tabular}{l|c|cc|cc}
            \toprule
                \multicolumn{2}{c|}{Subset}& \multicolumn{2}{c|}{SAS}& \multicolumn{2}{c}{DAS}\\
            \midrule
                 \multicolumn{2}{c|}{\multirow{2}{*}{\tabincell{c}{Perturbation\\(Total)}}}&\small{Change}&\small{Removal}&\small{Change}&\small{Removal}\\

               \multicolumn{2}{c|}{} &(4,004)&(3,986)&(4)&(22)\\
            \midrule
                 Metric&SE& \multicolumn{2}{c|}{$\Delta$ accuracy} & \multicolumn{2}{c}{accuracy}\\
                 \midrule
                 GPT-3.5&$\pm$0.49&\textbf{+0.47}&\textbf{-0.23}&25.00&0\\
                 GPT-4&$\pm$0.22&+0.38&	-1.20&\underline{50.00}&0\\
                 Gemini&$\pm$0.35&\textbf{-0.35}&-0.85&0&0\\
                 Claude3 &$\pm$0.22&+0.25&-0.40&25.00&\underline{4.55}\\
                 LLaMA2&$\pm$0.41&+0.47&-0.45&0&0\\
                 \bottomrule
            \end{tabular}}
            \caption{The sensitivity results on age.}
            \label{tab:age}
        \end{table}

    For the \textit{Different Answer Subset}, due to the limited total number of related samples, 
    GPT-4 can correctly answer 2 questions from 4 questions (i.e., 50\%) under age change perturbation, while GPT-3.5 and Claude3 can only answer 1 question correctly with an accuracy of 25\%. 
    For age removal perturbation, Claude3 is the only model that can answer a question correctly from 22 questions (i.e., 4.55\%), while all the other models are unable to answer any questions correctly.
    Overall, GPT-4 shows the highest sensitivity to age changes among the LLM models, while Claude3 achieves the highest sensitivity to age removal.

    \subsection{Sensitivity Analysis on Symptom}  
    Table \ref{tab:symptom} shows the analysis results on symptom. 
    For the \textit{Same Answer Subset}, Gemini shows a drastic decrease in $\Delta$ accuracy, dropping by 4.69\% for symptom changes and 2.34\% for removals. This shows its high sensitivity to symptom-related changes that should not influence the final diagnosis. 
    In contrast, GPT-3.5 demonstrates the most stable performance, with a small decrease of 0.23\% in $\Delta$ accuracy for symptom changes, and an increase of 0.24\% after symptom removal, both falling within its standard error range. The analysis results have suggested that GPT-3.5 is less sensitive to symptom changes that are unrelated to diagnosis results compared to other models.
    
        \begin{table}[h]
            \centering
            \setlength{\tabcolsep}{0.8mm}{
            \begin{tabular}{l|c|cc|cc}
            \toprule
                \multicolumn{2}{c|}{Subset}& \multicolumn{2}{c|}{SAS}& \multicolumn{2}{c}{DAS}\\
            \midrule
                 \multicolumn{2}{c|}{\multirow{2}{*}{\tabincell{c}{Perturbation\\(Total)}}}&\small{Change}&\small{Removal}&\small{Change}&\small{Removal}\\

               \multicolumn{2}{c|}{} &(542)&(2,395)&(2,921)&(1,068)\\
            \midrule
                 Metric&SE& \multicolumn{2}{c|}{$\Delta$ accuracy} & \multicolumn{2}{c}{accuracy}\\
                 \midrule
                 GPT-3.5&$\pm$0.49&\textbf{-0.23}&\textbf{+0.24}&0.10&1.97\\
                 GPT-4&$\pm$0.22&+1.88&	-0.70&\underline{3.08}&3.65\\
                 Gemini&$\pm$0.35&-4.69&-2.34&1.37&\underline{3.75}\\
                 Claude3&$\pm$0.22&-1.41&-1.88&0.86&0.94\\
                 LLaMA2&$\pm$0.41&-5.87&-4.93&0.03&0.47\\
                 \bottomrule
            \end{tabular}}
            \caption{The sensitivity results on symptom.}
            \label{tab:symptom}
        \end{table}
        
    For the \textit{Different Answer Subset}, GPT-4 outperforms other models, correctly answering 90 questions from 2,921 questions (i.e., 3.08\%) for symptom change.
    In terms of symptom removal, Gemini can correctly answer 40 questions from 1068 questions (i.e., 3.75\%), slightly surpassing GPT-4's accuracy of 3.65\%.    
    The results indicate that GPT-4 is most sensitive to symptom changes, while Gemini is most sensitive to symptom removal. Other models, such as LLaMA2 and Claude3, show difficulty in achieving good performance after perturbations.

    \subsection{Sensitivity Analysis on Checkup}  

    Table \ref{tab:checkup} shows the analysis results on checkup. 
    For the \textit{Same Answer Subset}, GPT-4 exhibits a slight increase of 0.37\% in $\Delta$ accuracy for checkup changes and a good performance for checkup removals. This suggests that GPT-4 achieves low sensitivity to irrelevant checkup perturbations. 
    In contrast, GPT-3.5 and Claude3 experience more noticeable accuracy changes, particularly GPT-3.5, which shows a decrease of 4.26\% for checkup changes and 5.90\% for removals in $\Delta$ accuracy, indicating high sensitivity to information that should not impact the diagnosis results.

     \begin{table}[h]
            \centering
            \setlength{\tabcolsep}{0.8mm}{
            \begin{tabular}{l|c|cc|cc}
            \toprule
                \multicolumn{2}{c|}{Subset}& \multicolumn{2}{c|}{SAS}& \multicolumn{2}{c}{DAS}\\
            \midrule
                 \multicolumn{2}{c|}{\multirow{2}{*}{\tabincell{c}{Perturbation\\(Total)}}}&\small{Change}&\small{Removal}&\small{Change}&\small{Removal}\\

               \multicolumn{2}{c|}{} &(1,719)&(3,028)&(1,720)&(411)\\
            \midrule
                Metric&SE& \multicolumn{2}{c|}{$\Delta$ accuracy} & \multicolumn{2}{c}{accuracy}\\
                 \midrule
                 GPT-3.5&$\pm$0.49&-4.26&-5.90&0.52&4.14\\
                  GPT-4&$\pm$0.22&	\textbf{0}&	+0.37&\underline{4.13}&\underline{5.25}\\
                 Gemini&$\pm$0.35&-2.50&-2.68&1.74&2.19\\
                 Claude3&$\pm$0.22&-1.83&-0.31&0.87&1.70\\
                 LLaMA2&$\pm$0.41&\textbf{-0.06}&-1.76&0.23&0.73\\
                 \bottomrule
            \end{tabular}}
            \caption{The sensitivity results on checkup.}
            \label{tab:checkup}
        \end{table}
        
    For the \textit{Different Answer Subset}, GPT-4 once again stands out, correctly answering 71 and 22 questions from 1720 checkup change questions and 411 checkup removal questions, with an accuracy of 3.08\% and 3.65\%, respectively.
    The results demonstrate GPT-4's ability to accurately detect significant checkup changes, suggesting it maintains the high sensitivity in scenarios where the diagnosis is affected by checkup changes.

    \section{Discussion}
    
\begin{figure*}[t]
        \centering
        \includegraphics[scale=0.4]{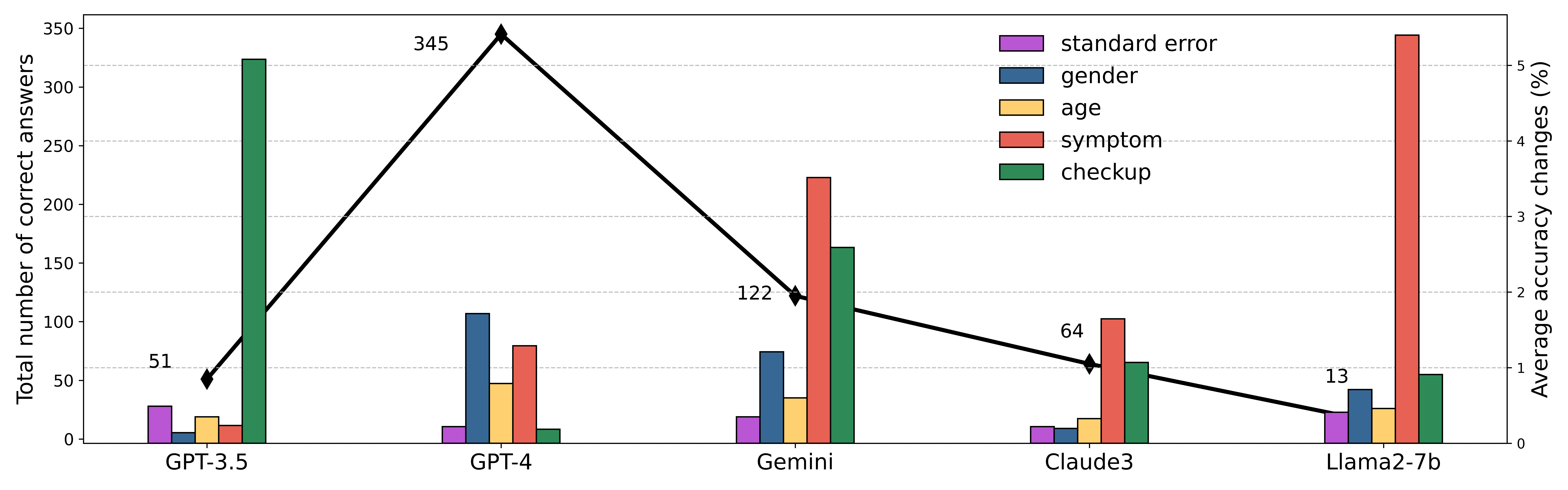}
        \caption{The overall sensitivity performance of five LLMs. The bar chart shows the average difference in accuracy on the Same Answer Subset, and the line chart shows the total correct answers provided by the LLMs on the Different Answer Subset.}
        \label{fig:fluctuation}
    \end{figure*}

    Figure \ref{fig:fluctuation} shows the overall sensitivity performance of five LLMs. 
    The bar chart illustrates the average $\Delta$ accuracy under two perturbations for each model on the \textit{Same Answer Subset} across the four kinds of key medical information.
    Meanwhile, the line chart depicts the total number of correct answers provided by the LLMs in the \textit{Different Answer Subset}.  
    
    From the perspective of whether LLMs have low sensitivity to questions in the \textit{Same Answer Subset}, GPT-3.5 is less sensitive to gender, age and symptom information, while GPT-4 is less sensitive to checkup.

    From the perspective of whether LLMs have high sensitivity to questions in the \textit{Different Answer Subset}, GPT-4 stands out with the highest number of correct answers (345), while Gemini follows with 122 correct answers.

    Overall, The results demonstrate GPT-4's high sensitivity to changes in key medical information that can alter diagnosis results, while maintaining moderate sensitivity to changes that do not affect diagnosis results.
    Gemini ranks as the second-best model in terms of sensitivity for the \textit{Different Answer Subset}; however, it also shows high sensitivity to less important changes, particularly symptom-related changes, which negatively impacts the robustness of the LLM. In contrast, Claude3 shows low sensitivity to the less important changes, but it also shows low sensitivity to the perturbations that affect diagnosis results, suggesting it may pay less attention to details. 
    
    However, even though GPT-4 has the highest sensitivity to important changes in key medical information among all models, demonstrating its potential as a medical diagnosis AI assistant, there remains a significant gap in its reliability when applied to real-world scenarios, as the accuracy only achieves 5.28\% (totally answering 345 questions from 6,528 questions correctly). Not to mention, GPT-4 is the most sensitive to gender perturbations among five LLMs. In conclusion, there is still a long way to go in reducing the sensitivity of LLMs to unimportant perturbations while enhancing their sensitivity to crucial ones.

\section{Conclusion}
    In this paper, we investigated the sensitivity of large language models (LLMs) to key medical information in clinical diagnostic decision-making. Our work is the first to explore the limitations of LLM's sensitivity from a clinical perspective. 
    First, we proposed a LLM sensitivity evaluation framework, within which we introduced four types of key medical information and designed two kinds of perturbation strategies. Based on this framework, we constructed a diagnosis-related dataset, DiagnosisQA, along with eight derived datasets. 
    We then evaluated five state-of-the-art LLMs on these datasets. The evaluation results reveal the limitations in large language model's ability to effectively capture key medical information. Most LLMs exhibit poor sensitivity to the key medical information. 
    Although large language models can achieve good performance on medical benchmarks, they still have substantial limitations in their sensitivity to key information in clinical diagnosis.

\section{Limitation}
    
    The dataset employed in the experiment was refined from MedQA. While these case-based QA questions closely resemble actual Electronic Medical Records (EMRs) used in clinical diagnosis, there are notable differences: (1) the information within the QA questions has been expert-validated and is highly relevant to problem-solving, whereas clinical EMRs in practice are generally longer and more comprehensive; (2) QA questions cover a limited number of diagnoses, leaving many common diagnoses uncovered, and the sensitivity of uncovered diagnoses is not investigated. Consequently, this presents a significant challenge for large language models. When applying LLMs to clinical practice, there remains a gap compared to our investigated results.

    The decision-making process in clinical diagnosis is highly sensitive to numerous crucial factors. In this context, we have only discussed four key medical information elements: gender, age, symptoms, and checkup results. However, there are other aspects not covered, such as family medical history. All of this information plays a pivotal role in guiding doctors to make accurate diagnosis.

\section*{Acknowledgments}
This research was supported by Noncommunicable Chronic Diseases-National Science and Technology Major Project (2023ZD0506501), National Key R\&D Program of China (2021YFC2500803) and Program of China Scholarship Council (Project ID:202306470061).

\bibliography{custom}

\appendix

    \section{The Impact of Prompts} 
        % \begin{table}[t]
        %     \centering
        %     \begin{tabular}{l|ccc}
        %         \toprule
        %          LLM& Accuracy &Recall &F1    \\
        %         \midrule
        %         GPT-3.5-Turbo& 57.20&46.08& 58.70\\
        %         GPT-4-Turbo&\textbf{76.49}&\textbf{63.22}& \textbf{71.85}\\
        %         Gemini& 58.39&48.62&61.31 \\ 
        %         LLaMA2-7b&2.27&1.78&3.48 \\
        %         Claude3-haiku&61.77&50.64&62.81 \\
        %          \bottomrule
        %     \end{tabular}
        %     \caption{The impact of prompts on the experimental results (\%) of LLMs on the DiagnosisQA dataset. The results of GPT-3.5-Turbo are the average of two trials, and the rest are in one-time experiment.} 
        %     \label{tab:diagnosisQA-withoutPrompt}
        % \end{table}

\begin{figure}[h]
            \centering
            \includegraphics[scale=0.44]{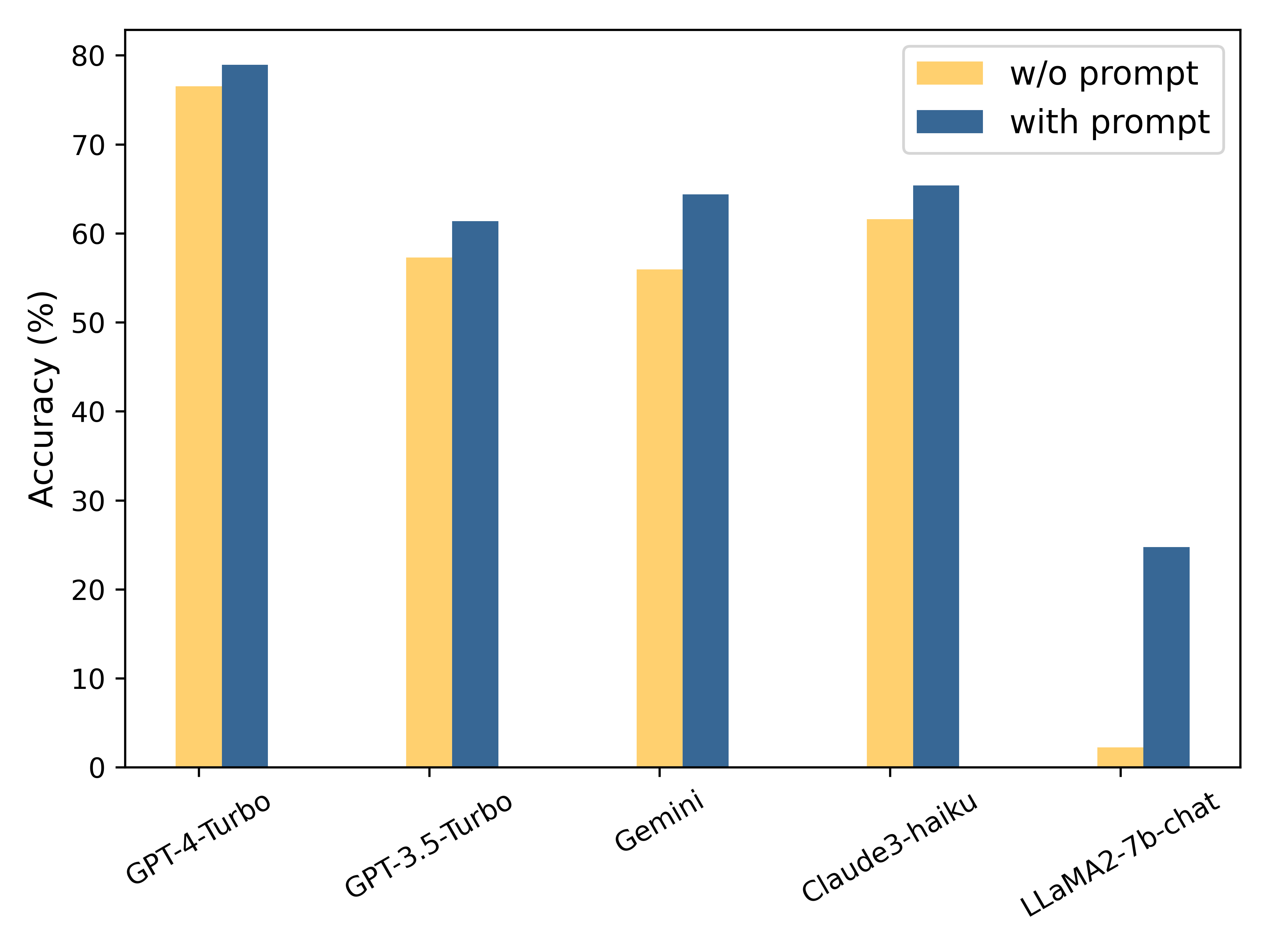}
            \caption{The impact of prompts on the performance of LLMs on the DiagnosisQA dataset.}
            \label{fig:diagnosisQA-withoutPrompt}
        \end{figure}

        To investigate the effects of prompts, we conduct a comparative experiment on the DiagnosisQA with empty prompt in the model input, but keeping the fields on Instruction, Question and Options. The results are shown in Figure \ref{fig:diagnosisQA-withoutPrompt}. 
        
        Overall, the use of prompts leads to a general improvement in the performance of all LLMs, indicating that prompts play a vital role in enhancing the understanding capability of LLMs. By incorporating elements such as the role, task and requirements.
        Specifically, for those LLMs with good performance, prompts bring a certain improvement. However, for the model with smaller parameters, i.e. LLaMA-7b, its performance drops sharply without prompts. Without prompts, LLaMA-7b can hardly understand the model input.

    \section{Case Study}

    % The patient's gender and age are crucial for clinical diagnosis. Some diseases may have varying incidence rates, symptom presentations, and treatment approaches between different genders and age groups. 
    % To demonstrate the impact of key medical information on diagnosis, we give an example from DiagnosisQA, illustrated in Table \ref{tab:case}. 
    
    % The age removal perturbation is applied to the question. 
    % In this example, GPT-3.5 does not pay particular attention to the age information, so fails to provide the correct answer even before introducing any perturbation.

    To demonstrate the impact of key medical information on diagnosis, we give an example from DiagnosisQA, illustrated in Table \ref{tab:appendix-case-human}. 
    % Table \ref{tab:appendix-case-human} provides another case study. 
    The gender change perturbation is applied to the question.
    The original question involves a female patient, and the gender change perturbation causes a conflict between the patient's gender and situation. Consequently, the correct answer becomes "The question contains inconsistencies". In the evaluation, we find that both GPT-4 and Gemini select option F and note the conflict between the male patient and the described symptoms and candidate diseases in their explanations. They also recommended patient to take further examinations. However, GPT-3.5 and Claude3 did not recognize this issue.

    \begin{table*}[htbp]
        \centering
        \begin{tabular}{l|c|c}
            \toprule
            & Original & With Gender Change Perturbation\\
            \midrule
             Question&
             \tabincell{l}{A 14-year-old \underline{girl} is brought to the \\pediatrician by her mother. The \underline{girl}'s \\mother states that \underline{she} began having her \\period 6 months ago. The patient states \\that after \underline{she} first period \underline{she} has had a \\period every 10 to 40 days. \underline{Her} menses \\have ranged from very light flow to in-\\tense and severe symptoms. Otherwise, \\the patient is doing well in school, is on \\the track team, and has a new \underline{boyfriend}. \\ \underline{Her} temperature is 98.1°F (36.7°C), \\blood pressure is 97/58 mmHg, pulse is \\90/min, respirations are 14/min, and \\oxygen saturation is 99\% on room air. \\Physical exam demonstrates an obese \\ \underline{girl} but is otherwise within normal limits. \\Which of the following is the most \\likely diagnosis?}
             & 
             \tabincell{l}{A 14-year-old \underline{boy} is brought to the \\pediatrician by his mother. The \underline{boy}'s \\mother states that \underline{he} began having his \\period 6 months ago. The patient states \\that after \underline{his} first period \underline{he} has had a \\period every 10 to 40 days. \underline{His} menses \\have ranged from very light flow to in-\\tense and severe symptoms. Otherwise, \\the patient is doing well in school, is on \\the track team, and has a new \underline{girlfriend}. \\ \underline{His} temperature is 98.1°F (36.7°C), \\blood pressure is 97/58 mmHg, pulse is \\90/min, respirations are 14/min, and \\oxygen saturation is 99\% on room air. \\Physical exam demonstrates an obese \\ \underline{boy} but is otherwise within normal limits. \\Which of the following is the most \\likely diagnosis?} \\
             \hline
             Options& \multicolumn{2}{c}{\tabincell{l}{
             A.\quad Normal development \quad \textbf{[Correct Answer Before Perturbation]} \\
             B.\quad Polycystic ovarian syndrome \\
             C.\quad Pregnancy \\
             D.\quad Premenstrual dysphoric disorder \\
             E.\quad Premenstrual tension \\
             F.\quad The question contains inconsistencies \quad \textbf{[Correct Answer After Perturbation]}}}\\
             \hline
             \tabincell{c}{GPT-4}& \multicolumn{2}{c}{\tabincell{l}{
             F. The scenario describes a 14-year-old boy experiencing menstrual periods, which\\ 
             is \textbf{physiologically impossible as males do not have the reproductive organs ne-}\\ \textbf{cessary for menstruation}. The presence of 'periods' in a male patient suggests a\\
             different underlying issue that may be related to hormonal imbalances, endocrine\\
             disorders, or other medical conditions that need to be investigated further. \textbf{None of}\\ \textbf{the provided options (A-E) are applicable to a male patient} experiencing these \\symptoms.\quad \CheckmarkBold}}\\
             \hline
            GPT-3.5&\multicolumn{2}{c}{\tabincell{l}{
            B. The most likely diagnosis in this case is polycystic ovarian syndrome (PCOS). \\
            PCOS is a hormonal disorder that affects women of reproductive age. It is charac-\\ terized by irregular menstrual periods, which can range from very light flow to \\ intense and severe symptoms. Other common symptoms of PCOS include obesity. \\
            The patient's age, irregular periods, and obesity are all suggestive of PCOS. The\\ other options are less likely based on the given information.\quad \XSolidBrush
            }}\\
            \hline
            Gemini&\multicolumn{2}{c}{\tabincell{l}{
            F. \textbf{The question contains inconsistencies.} The patient is a 14-year-old boy who \\has been having menstrual periods for 6 months. This is not normal development, \\\textbf{as boys do not typically have periods}. Polycystic ovarian syndrome, pregnancy, \\premenstrual dysphoric disorder, and premenstrual tension are all conditions that \\can cause irregular periods in women, but \textbf{they are not typically seen in boys.} The\\ 
            patient should be evaluated further to determine the cause of irregular periods.\quad \CheckmarkBold}}\\
            \hline
            Claude3&\multicolumn{2}{c}{\tabincell{l}{
            B. Polycystic ovarian syndrome (PCOS) is a common endocrine disorder that affects \\
            women of reproductive age. It is characterized by irregular periods, excess androgen \\ levels, and the presence of cysts on the ovaries. PCOS can cause a variety
            of symp-\\toms, including acne, hirsutism, weight gain, and infertility. In this case, the patient \\is a 14-year-old boy who has been having irregular periods for the past 6 months. He \\is also obese, which is a common symptom of PCOS.\quad \XSolidBrush
            
            }}\\
            \bottomrule
        \end{tabular}
        \caption{A case study of the perturbed questions with the answer of "The question contains inconsistencies".
        }
        \label{tab:appendix-case-human}
    \end{table*}

\end{document}